\documentclass[10pt,twocolumn,letterpaper]{article}

\usepackage{cvpr}
\usepackage{times}
\usepackage{epsfig}
\usepackage{graphicx}
\usepackage{amsmath}
\usepackage{amssymb}
\usepackage{capt-of,etoolbox}
\usepackage{wrapfig}
\usepackage[title]{appendix}
\usepackage{authblk}

\usepackage{soul}
\usepackage{xcolor}

\def\figurePath{images/}
\def\myfigure#1#2{\begin{figure}[t]\centering\includegraphics*[width = \linewidth]{\figurePath#1}\caption{#2}\label{fig:#1}\vspace{-5pt}\end{figure}}
\def\mycfigure#1#2{\begin{figure*}[t]\centering\includegraphics*[clip, width = \linewidth]{\figurePath#1}\caption{#2}\label{fig:#1}\vspace{-6pt}\end{figure*}}

\def\mysection#1#2{\section{#1}\label{sec:#2}}

\newcommand{\refSec}[1]{Sec.~\ref{sec:#1}}
\newcommand{\refFig}[1]{Fig.~\ref{fig:#1}}

\newcommand{\refTbl}[1]{Table~\ref{tbl:#1}}

\definecolor{unsurecolor}{rgb}{1,.85,.7}
\definecolor{changedcolor}{rgb}{.85,1,.7}

\soulregister\ref7
\soulregister\cite7
\soulregister\citeetal7
\soulregister\etal0
\soulregister\eg0
\soulregister\scene1
\soulregister\refTbl1

\DeclareGraphicsExtensions{.pdf,.ai,.psd,.jpg}
\usepackage[pagebackref=true,breaklinks=true,letterpaper=true,colorlinks,bookmarks=false]{hyperref}

\cvprfinalcopy 

\def\cvprPaperID{310} 

\ifcvprfinal\fi
\begin{document}

\title{Soccer on Your Tabletop}

\makeatletter
\renewcommand\AB@affilsepx{, \protect\Affilfont}
\makeatother

\author[1]{Konstantinos Rematas}
\author[1,2]{Ira Kemelmacher-Shlizerman}
\author[1]{Brian Curless}
\author[1,3]{Steve Seitz}
\affil[1]{University of Washington}
\affil[2]{Facebook}
\affil[3]{Google}

\twocolumn[{%
\renewcommand\twocolumn[1][]{#1}%
\maketitle
\begin{center}
    \centering
    \includegraphics[width=.98\textwidth]{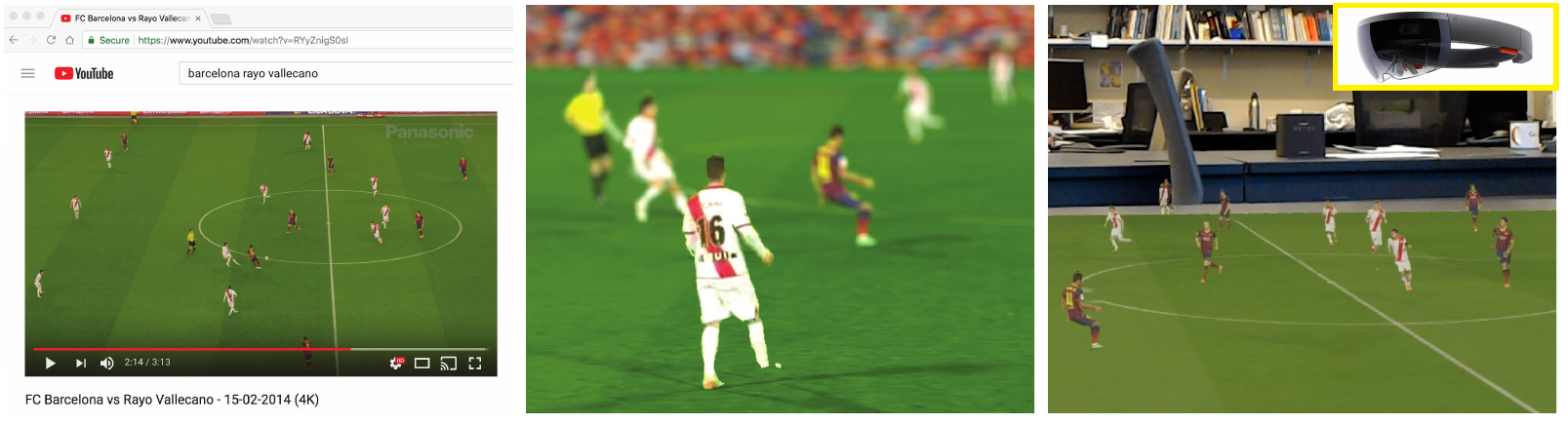}
    \captionof{figure}{From a YouTube video of a soccer game, our system outputs a dynamic 3D reconstruction of the game, that can be viewed interactively on your tabletop with an Augmented Reality device. The \textbf{supplementary video} demonstrates  the capabilities of the method.}
    \label{fig:teaser}
\end{center}%
}]

\newcommand{\icol}[1]{
  \left(\begin{smallmatrix}#1\end{smallmatrix}\right)%
}

\begin{abstract}
We present a system that transforms a monocular video of a soccer game into a moving 3D reconstruction, in which the players and field can be rendered interactively with a 3D viewer or through an Augmented Reality device.  At the heart of our paper is an approach to estimate the depth map of each player, using a CNN that is trained on 3D player data extracted from soccer video games.  We compare with state of the art body pose and depth estimation techniques, and show results on both synthetic ground truth benchmarks, and real YouTube soccer footage.
\end{abstract}

\mysection{Introduction}{intro}

Imagine watching a 3D hologram of a live soccer game on your living room table; you can walk around with an Augmented Reality device, watch the players from different viewpoints, and lean in to see the action up close.

One way to create such an experience is to equip the soccer field with many cameras, synchronize the cameras, and then reconstruct the field and players in 3D using multi-view geometry techniques. Approaches of that spirit were previously proposed in the literature \cite{Ger10,Germann10,Guillemaut2011} and even commercialized as  Replay's FreeD, and others \cite{freed}.   The results of multi-view methods are  impressive, however the requirement of physically instrumenting the field with many synchronized cameras limits their generality.  What if, instead, we could reconstruct any soccer game just from a single YouTube video?  This is the goal of this paper. 

There are numerous challenges in monocular reconstruction of a soccer game. We must estimate the camera pose relative to the field, detect and track each of the players, reconstruct their body shapes and poses, and render the combined reconstruction. 

We present the first end-to-end system (\refFig{overview_smallsize}) that accomplishes this goal (short of reconstructing the ball, which remains future work). In addition to the system, a key technical contribution of our paper is a novel method for player body depth map estimation from a single frame. Our approach is trained on meshes extracted from FIFA video games. Based on this data, a neural network estimates per pixel depth values of any new soccer player, comparing favorably to other state-of-the-art body depth and pose estimation techniques. 

We present results on 10 YouTube games of different teams.
Our results can be rendered using any 3D viewer, enabling free-viewpoint navigation from the side of the field recorded by the game camera.  We also implemented ``holographic'' Augmented Reality viewing with HoloLens, projected onto a tabletop. See the supplementary material for the AR video results and the 3D model of the game. 

\mycfigure{overview_smallsize}{Overview of our reconstruction method. From a YouTube video frame, we recover camera parameters using the field lines.  We then extract bounding boxes, poses, and trajectories (across multiple frames) to segment the players.  Using a deep network trained on video game data, we reconstruct per-player depth maps on the playing field, which we can render in a 3D viewer or on an AR device.}

\mysection{Related Work}{related}

\paragraph{Sports Analysis} Sports game analysis has been extensively investigated from the perspectives of image processing, computer vision, and computer graphics~\cite{Moeslund}, both for academic research and for industry applications. Understanding a sports game involves several steps, from field localization to player detection, tracking, segmentation, \etc. Most sports have a predefined area where the action is happening; therefore, it is essential to localize that area \wrt the camera. This can be done with manual correspondences and calibration based on, \eg, edges~\cite{Carr2012}, or fully automatically ~\cite{NamdarCVPR17}. In this work, we follow a field localization approach similar to~\cite{Carr2012}.

Sports reconstruction can be achieved using multiple cameras or specialized equipment, an approach that has been applied to  free viewpoint navigation and 3D replays of games. Products such as Intel FreeD~\cite{freed} produce new viewing experiences by incorporating data from multiple cameras. Similarly, having a multi-camera setup allows multiview stero methods~\cite{Guillemaut09,Guillemaut2011} for free viewpoint navigation~\cite{Grau2007ARF, Thomas2006, Grau2006AFV},  view interpolation based on player triangulation~\cite{Ger10} or view interpolation by representing players as billboards~\cite{Germann10}.   In this paper, we show that reliable reconstruction from {\em monocular} video is now becoming possible due to recent advances in people detection~\cite{Ren2015, Dai16}, tracking \cite{Milan2014}, pose estimation~\cite{Wei2016ConvolutionalPM, pishchulin16cvpr},  segmentation~\cite{he2017maskrcnn}, and deep learning networks.  In our framework, the input is broadcast video of a game, readily available on YouTube and other online media sites.

\vspace{-10pt}
\paragraph{Human Analysis} Recently, there has been enormous improvement in people analysis using deep learning. Person detection~\cite{Ren2015, Dai16} and pose estimation~\cite{Wei2016ConvolutionalPM, pishchulin16cvpr} provide robust building blocks for further analysis of images and video. Similarly, semantic segmentation can provide pixel-level predictions for a large number of classes~\cite{Yu2015MultiScaleCA,KunduCVPR16}. In our work, we use such predictions (bounding boxes from~\cite{Ren2015}, pose keypoints~\cite{Wei2016ConvolutionalPM}, and people segmentation~\cite{Yu2015MultiScaleCA}) as input steps towards a full system where the input is a single video sequence, and the output is a 3D model of the scene.

Analysis and reconstruction of people from depth sensors is an active area of research~\cite{shotton12, BogoICCV2015}, but the use of depth sensors in outdoor scenarios is limited because of the interference with abundant natural light. An alternative would be to use synthetic data~\cite{varol17b,JohnsonRoberson2017DrivingIT, Taylor2007OVVVUV, Shafaei2016PlayAL}, but these virtual worlds are far from our soccer scenario.
There is extensive work on depth estimation from images/videos of indoor~\cite{eigen14} and road~\cite{Godard17} scenes, but not explicitly for humans. Recently, the work of~\cite{varol17b} proposes a human part and depth estimation method trained on synthetic data. They fit a parametric human model~\cite{Loper2015SMPLAS} to motion capture data and use cloth textures to model appearance variability for arbitrary subjects and poses when constructing their dataset. In contrast, our approach takes advantage of the restricted soccer scenario for which we construct a dataset of depth map / image pairs of players in typical soccer clothing and body poses extracted from a high quality video game. Another approach that can indirectly infer depth for humans from 2D images is ~\cite{Bogo2016KeepIS}. This work estimates the pose and shape parameters of a 3D parametric shape model in order to fit the observed 2D pose estimation. However, the method relies on robust 2D poses, and the reconstructed shape does not fit to the players' clothing. We compare to both of these methods in the Experiments section. 

Multi-camera rigs are required for many motion capture and reconstruction methods~\cite{deAguiar2008, Starck2007SurfaceCF}. \cite{MustafaCVPR17} uses a CNN person segmentation per camera and fuses the estimations in 3D.  Body pose estimation from multiple cameras is used for outdoor motion capture in \cite{rhodin_ECCV2016, EEJTP15}. In the case of a single camera, motion capture can be obtained using 3D pose estimators~\cite{pavlakos2017volumetric, pavlakos2017harvesting, VNect_SIGGRAPH2017}. However, these methods provide the 3D position only for skeleton joints; estimating full human depth would require additional steps such as parametric shape fitting. We require only a single camera.

\mysection{Soccer player depth map estimation}{neuralnet}

A key component of our system is a method for estimating a depth map for a soccer player given only a single image of the player.  In this section, we describe how we train a deep network to perform this task.

\myfigure{fifa_data_small}{Training data: we extracted images and their corresponding depths while playing a FIFA game. We present several examples here visualized as depth maps and meshes.}

\subsection{Training data from FIFA video games}
State-of-the-art datasets for human shape modeling mostly focus on general representation of human bodies and aim at diversity of body shape and clothing \cite{Loper2015SMPLAS,varol17b}. Instead, to optimize for accuracy and performance in our problem, we  want a training dataset that focuses solely on soccer, where clothing, players' poses, camera views, and positions on the field are very constrained. Since our goal is to estimate a depth map given a single photo of a soccer player, the ideal training data would be image and depth map pairs of soccer players in various body poses and clothing, viewed from a typical soccer game camera.

The question is: how do we acquire such ideal data?   It turns out that while playing Electronic Arts FIFA games and intercepting the calls between the game engine and the GPU ~\cite{Richter_2016_ECCV, Richter_2017}, it is possible to extract depth maps from video game frames. 

In particular, we use RenderDoc \cite{renderdoc} to intercept the calls between the game engine and the GPU. FIFA, similar to most games, uses deferred shading during game play. Having access to the GPU calls enables capture of the depth and color buffers per frame\footnote{RenderDoc causes the game to freeze, essentially capturing 1 fps.}.  Once depth and color is captured for a given frame we process it to extract the players. 

The extracted color buffer is an RGB screen shot of the game, without the score and time counter overlays and the in-game indicators. 
The extracted depth buffer is in Normalized Device Coordinates (NDC), with values between 0 and 1. To get the world coordinates of the underlying scene we require the OpenGL camera matrices that were used for rendering. In our case,
these matrices were not directly accessible in RenderDoc, so we estimated them (see Appendix A in supplementary material).

Given the game camera parameters, we can convert the z-buffer from the NDC to 3D points in world coordinates. The result is a point cloud that includes the players, the ground, and portions of the stadium when it is visible. The field lies in the plane $y=0$.  To keep only the players, we remove everything that is outside of the soccer field boundaries and all points on the field (i.e., points with $y=0$). To separate the players from each other we use DBSCAN clustering~\cite{Ester1996} on their 3D locations. Finally, we project each player's 3D cluster to the image and recalculate the depth buffer with metric depth. Cropping the image and the depth buffer around the projected points gives us the image-depth pairs -- we extracted $12000$ of them -- for training a depth estimation network (\refFig{fifa_data_small}). Note that we use a player-centric depth estimation because we get more training data by breaking down each frame into 10-20 players, and it is easier for the network to learn individual player's configuration rather than whole-scene arrangements.

\subsection{Depth Estimation Neural Network}
\label{sec:depth_estimation}
Given the depth-image pairs extracted from the video game, we train a neural network to estimate depth for any new image of a soccer player.  Our approach follows the hourglass network model~\cite{Newell16,varol17b}: the input is processed by a sequence of hourglass modules -- a series of residual blocks that lower the input resolution and then upscale it -- and the output is depth estimates. 

Specifically, the input of the network is a $256\times256$ RGB image cropped around a player together with a segmentation mask for the player, resulting in a 4-channel input.  We experimented with training on no masks, ground truth masks, and estimated masks.  Using masks noticeably improved results.  In addition, we found that using estimated masks yielded better results than ground truth masks.  With estimated masks, the network learns the noise that occurs in player segmentation during testing, where no ground truth masks are available. To calculate the player's mask, we apply the person segmentation network of~\cite{Yu2015MultiScaleCA}, refined with a CRF~\cite{Krhenbhl2011EfficientII}. Note that our network is single-player-centric: if there are overlapping players in the input image, it will try to estimate the depth of the center one (that originally generated the cropped image) and assign the other players' pixels to the background.

The input is processed by a series of 8 hourglass modules and the output of the network is a $64\times64\times50$ volume, representing 49 quantized depths (as discrete classes) and 1~background class. The network was trained with cross entropy loss with batch size of 6 for 300 epochs with learning rate 0.0001 using the Adam~\cite{Kingma2014AdamAM} solver (see  details of the architecture in supplementary material). 

The depth parameterization is performed as follows: first, we estimate a virtual vertical plane passing through the middle of the player and calculate its depth \wrt the camera. Then, we find the distance in depth values between a player's point and the plane. The distance is quantized into 49 bins (1 bin at the plane, 24 bins in front, 24 bins behind) at a spacing of 0.02 meters, roughly covering 0.5 meters in front and in back of the plane (1 meter depth span). In this way, all of our training images have a common reference point. Later, during testing, we can apply these distance offsets to a player's bounding box after lifting it into 3D (see \refSec{mesh-generation}).

\mysection{Reconstructing the Game}{system}
In this section we describe our full pipeline for 3D reconstruction from a soccer video clip. 

\subsection{Camera Pose Estimation}
\label{sec:localization}

The first step is to estimate the per-frame parameters of the real game camera.
Because soccer fields have specific dimensions and structure according to the rules of FIFA, we can estimate the camera parameters by aligning the image with a synthetic planar field template. We set the world origin to coincide with the center of the synthetic soccer field which lies in the $y=0$ plane.

The most consistent features on the field of play are the field lines (\eg, sidelines, penalty box around the goal).
Thus, we extract edge points $\mathcal{E}$ for each frame to localize those features.  We can  solve for the camera parameters $\mathbf{w}$ (focal length, rotation and translation) that align rendered synthetic field lines with the extracted edge points.  In particular, we first construct a distance map $\mathcal{D}$ that, for each pixel in the original frame, stores the squared distance to the nearest point in $\mathcal{E}$.  Then,
for projection $\mathcal{T}(p; \mathbf{w})$ that maps the visible 3D line points $p$ to the image,  we minimize:
\begin{equation}
\min_{\mathbf{w}} \sum_p \mathcal{D}\left(\mathcal{T}(p; \mathbf{w})\right),
\end{equation}
\ie, the sum of squared distances between the projected synthetic field points and the nearest edge points in $\mathcal{E}$.

This process is highly dependent on the the quality of the edge points $\mathcal{E}$ and the camera initialization. We use structured forests~\cite{DollarICCV13edges} for line detection. We additionally remove edges that belong to people by applying a person segmentation network~\cite{Yu2015MultiScaleCA}. To initialize the camera fitting, we provide 4 manual correspondences in the first frame, and further solve for the camera pose in each successive frame using the previous frame as initialization.

\subsection{Player Detection and Tracking}

The first step of the video analysis is to detect the players in every frame. While detecting soccer players may seem straightforward due to the relatively uniform background, most state-of-the-art person detectors still have difficulty when, \eg, players from the same team occlude each other or the players are too small.

We start with a set of bounding boxes obtained with~\cite{ren2015faster}. Next, we refine the initial bounding boxes based on pose information using the detected keypoints/skeletons from~\cite{Wei2016ConvolutionalPM}. We observed that the estimated poses can better separate the players than just the bounding boxes, and the pose keypoints can be effectively used for tracking the players across frames.

Finally, we generate tracks over the sequence based on the refined bounding boxes. Every track has a starting and ending location in the video sequence. The distance between two tracks A and B is defined as the 2D Euclidean distance between the ending location of track A and starting location of track B, assuming track B starts at a later frame than track A and their frame difference is smaller than a threshold (detailed parameters are described in supplementary material). 
We follow a  greedy merging strategy. We start by considering all detected neck keypoints (we found this keypoint to be the most reliable to associate with a particular player) from all frames as separate tracks and we calculate their pairwise distances. Two tracks are merged if their distance is below a threshold, and we continue until there are no tracks to merge. This step associates 
every player with a set of bounding boxes and poses across frames. This information is essential for the later processing of the players, namely the temporal segmentation, depth estimation and better placement in 3D. \refFig{overview_smallsize} shows the steps of detection, pose estimation, and tracking.

\subsection{Temporal Instance Segmentation}
For every tracked player we need to estimate its segmentation mask to be used in the depth estimation network. A straightforward approach is to apply at each frame a person segmentation method~\cite{Yu2015MultiScaleCA}, refined with a dense CRF~\cite{Krhenbhl2011EfficientII} as we did for training. This can work well for the unoccluded players, but in the case of overlap, the network estimates are confused. Although there are training samples with occlusion, their number is not sufficient for the network to estimate the depth of one player (\eg the one closer to the center) and assign the rest to the background. For this reason, we ``help'' the depth estimation network by providing a segmentation mask where the tracked player is the foreground and the field, stadium and other players are background (this is similar to the instance segmentation problem~\cite{he2017maskrcnn, li2016fully}, but in a 1-vs-all scenario). 

To estimate the pixels that belong to a particular player $T$, we rely both on the semantic segmentation and on the pose estimation from the previous step. First, for every pixel $p$, we aim to find the continuous variable $o_p$ that indicates its association to the player $T$, background or other players by minimizing the energy~\cite{Levin2004, Krishnan2013,KunduCVPR16}:
\begin{equation}
E = \sum_{p} (o_p - \sum_{q\in N(p)} w_{pq}o_q),
\end{equation}
where $N(p)$ is the spatial neighborhood of pixel $p$, and $w_{pq}$ is the affinity between pixels $p$ and $q$ based on the color image $I$ and edge image $G$: $\exp(-||I_p-I_q||^2)*\exp(-G_p^2)$. Several pixels can be used as anchors for the optimization: a) the pixels $s$ that belong to the tracked player skeleton will have $o_s=0$, b) other players' skeleton pixels $r$ have $o_r=2$ and c) pixels $b$ with high background probability have $o_b=1$. By thresholding the optimized $o_p$ values (we use 0.5 for our experiments) we generate the player's mask $M_o$. This mask performs well in separating the main player from other players, but tends to include some background as well. 

To better segment out the background, we estimate an additional mask $M_{\rm \it CNN}$ as follows. We construct a video volume containing the player in a block of 15 frames, with the player's per-frame locations translated to align the neck keypoint across frames. We solve a dense CRF~\cite{Krhenbhl2011EfficientII} over the volume to obtain $M_{\rm \it CNN}$ for every frame in the block. The unary potentials come from the person segmentation network of~\cite{Yu2015MultiScaleCA}.
The pairwise potentials are modeled as Gaussian kernels in a $D$-dimensional feature space, with the features $f\in D$ consisting of the $rgb$ colors, the $xy$ locations, and $t$ the time stamp. $M_{\rm \it CNN}$ better segments out the background, but tends to include other players.  Thus, our final segmentation mask is the product $M_{\rm \it final} = M_o M_{\rm \it CNN}$. In the inset image, we show an occluded player, the optimized variables $o_p$, our masks, and the instance segmentation from another state-of-the-art method~\cite{li2016fully} (see supplementary for additional results). For the $o_p$ visualization, $o_s$ is yellow, $o_r$ is blue, and $o_b$ is magenta.
\begin{figure}[h]
\vspace{-5pt}
\centering
\includegraphics[width=0.95\columnwidth]{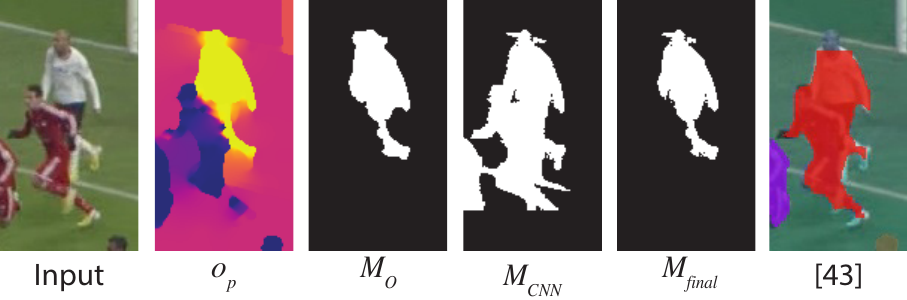}
\vspace{-10pt}
\end{figure}

\subsection{Mesh Generation}
\label{sec:mesh-generation}
The foreground mask from the previous step, together with the original cropped image are fed to the network described in \ref{sec:depth_estimation}. The output of the network is per-pixel, quantized signed distances between the player's surface and a virtual plane \wrt the camera. To obtain a metric depth map we first lift the bounding box of the player into 3D, creating a billboard (we assume that the bottom pixel of the player lies on the ground). We then apply the distance offsets output by the network to the 3D billboard to obtain the desired depth map. 

The depth map is then unprojected to world coordinates using the camera parameters, generating the player's pointcloud in 3D. Each pixel corresponds to a 3D point and we use pixel connectivity to establish faces. We texture-map the mesh with the input image. Depending on the application, the mesh can be further simplified with mesh decimation to reduce the file size for deployment in an AR device.

\subsection{Trajectories in 3D}

Due to imprecise camera calibration and bounding box localization, the 3D placement of players can ``jitter'' from frame to frame.  To address this problem, we smooth the 3D trajectories of the players.  In particular, once we estimate the player's position in the 3D field, we calculate the center of the mesh (mean of the player's vertices) and solve for its optimized 3D trajectory $\mathbf{X}\in \mathbb{R}^{N \times 3}$~\cite{Milan2014} by minimizing:
\begin{equation}
E = \sum_{t\in M}||\mathbf{X}_t-D_t||^2 + \sum_{t=1}^{N-1} || \mathbf{X}_{t-1} - 2\mathbf{X}_{t}+\mathbf{X}_{t+1}||^2
\end{equation}
where $N$ is the number of frames and $M$ is the set of timestamps when a detection occurs. $D_t$ corresponds to the center of the lifted bounding box in 3D at time $t$. The first term of the objective ensures that the estimated trajectory will be close to the original detections, and the second term encourages second order temporal smoothness.

\mysection{Experiments}{experiments}
All videos were processed in a single desktop with an i7 processor, 32 GB of RAM and a GTX 1080 with 6GB of memory. The full (unoptimized) pipeline takes approximately 15 seconds for a typical 4K frame with 15 players.
\mycfigure{synth_results_smallsize}{Results on the synthetic dataset and comparison to state of the art and ground truth, visualized as depth maps and 3D meshes.
Our method infers more accurate and complete depth estimates, which result in better mesh reconstructions.}

\vspace{-10pt}
\paragraph{Synthetic Evaluation} 
We quantitatively evaluate our approach and several others using a held-out dataset from FIFA video game captures. The dataset was created in the same way as the training data (\refSec{neuralnet}) and contains 32 rgb-depth pairs of images, containing  450 players. We use the scale invariant root mean square error ($st$-RMSE)~\cite{varol17b,eigen14} to measure the deviation of the estimated depth values of foreground pixels from the ground truth. In this way we compensate for any scale/translation ambiguity along the camera's $z$-axis. We additionally report segmentation accuracy results  using the intersection-over-union (IoU) metric.

We compare with three different approaches: a) non human-specific depth estimation~\cite{Chen2016}, b)  human-specific depth estimation~\cite{varol17b}, and c) fitting a parametric human shape model to 2D pose estimations~\cite{Bogo2016KeepIS}. For all of these methods, we use their publicly available code.

The input for all methods are cropped images containing soccer players. We apply the person detection and pose estimation steps, as described in~\refSec{system}, to the original video game images in order to find the same set of players for all methods (resulting in 432 player-depth pairs). For each detection, we crop the area around the player to use as a test image, and we get its corresponding ground truth depth for evaluation. In addition, we lift its bounding box in 3D to get the location of the player in the field and to use it for our depth estimation method (note that the bounding box is not always tight around the player, resulting in some displacement across the camera's z-axis).

The cropped images come from a larger frame with known camera parameters; therefore, the depth estimates can be placed back in the original camera's (initially empty) depth buffer.
Since the depth estimates from the different methods depend on the camera settings that each method used during training, it is necessary to use scale/translation invariance metrics. 
In addition, we transform the output of \cite{varol17b} into world units by multiplying by their quantization factor (0.045m). Note that our estimates are also in world units, since we use the exact dimensions of the field for camera calibration. For~\cite{Bogo2016KeepIS}, we modify their code to use the same 2D pose estimates used in our pipeline~\cite{Wei2016ConvolutionalPM} and we provide the camera parameters and the estimated 3D location of the player.
Table~\ref{table:results} summarizes the quantitative results for depth estimation and player segmentation. Our method outperforms the alternatives both in terms of depth error and player coverage. This result highlights the benefit of having a training set tailored to a specific scenario. 
\begin{table}[h]
\centering
\begin{tabular}{ lll }
& st-RMSE & IoU\\
  \hline		
 	Non-human training~\cite{Chen2016} & 0.92 & - \\
 	Non-soccer training~\cite{varol17b} & 0.16 & 0.41\\
	Parametric Shape~\cite{Bogo2016KeepIS} & 0.14 & 0.61 \\
	Ours & \textbf{0.06} & \textbf{0.86} \\
\end{tabular}
\caption{Depth estimation ($st$-RMSE) and player segmentation (IoU) comparison for different approaches.}\label{table:results}
\end{table}

\myfigure{overlay_smallsize}{Reconstructed mesh comparison.}

In \refFig{synth_results_smallsize} we show qualitative results on the synthetic dataset, both as depth maps and meshes. Our approach is closer to the ground truth posture and mesh shape, since our network has been trained on people with soccer outfits and common soccer poses (\refFig{overlay_smallsize}). 

The method of~\cite{varol17b} assigned a large number of foreground pixels to the background. One reason is that their training data aims to capture general human appearance against cluttered backgrounds, unlike what is found in typical soccer images. Moreover, the parametric shape model~\cite{Loper2015SMPLAS} that is used in \cite{varol17b, Bogo2016KeepIS} is based on scans of humans with shapes and poses not necessarily observed in soccer games. Trying to fit such a model to soccer data may result in shapes/poses that are not representative of soccer players. 
In addition, the parametric shape model is trained on subjects wearing little clothing, resulting in ``naked'' reconstructions.

\vspace{-10pt}
\paragraph{YouTube videos} 
We evaluate our approach on a collection of soccer videos 
downloaded from YouTube with 4K resolution. The initial sequences were trimmed to 10 video clips shot from the main game camera. Each sequence is 150-300 frames and contains various game highlights (\eg, passing, shooting, \etc) for different teams and with varying numbers of players per clip. The videos also contain typical imaging artifacts, such as chromatic aberration and motion blur, and compression artifacts.

\myfigure{reals_smallsize}{Results on real images from YouTube videos.}

\refFig{reals_smallsize} shows the depth maps of different methods on real examples. Similar to the synthetic experiment, the non-human and non-soccer methods perform poorly. The method of~\cite{Bogo2016KeepIS} correctly places the projections of the pose keypoints in 2D, but the estimated 3D pose and shape are often different from what is seen in the images. Moreover, the projected parametric shape does not always correctly cover the player pixels (also due to the lack of clothing), leading to incorrect texturing~(\refFig{real_meshes_smallsize}). With our method, while we do not obtain full 3D models as in~\cite{Bogo2016KeepIS}, the visible surfaces are modeled properly (\eg the player's shorts). Also, after correctly texturing our 3D model, the quantization artifacts from the depth estimation are no longer evident.  In principle, the full 3D models produced by~\cite{Bogo2016KeepIS} could enable viewing a player from a wide range of viewpoints (unlike our depth maps); however, they will lack correct texture for unseen portions in a given frame, a problem that would require substantial additional work to address.

\myfigure{real_meshes_smallsize}{Reconstructed meshes and texturing. }

\myfigure{kth_small}{Our depth estimates are consistent for the same pose but different viewpoints (colors indicate viewpoint).}
\myfigure{time_consistency_smallsize}{Depth estimation consistency for consecutive frames from 25 fps video, even the training data was captured at 1 fps.}

\begin{figure*}[h]
  \includegraphics[width=\textwidth]{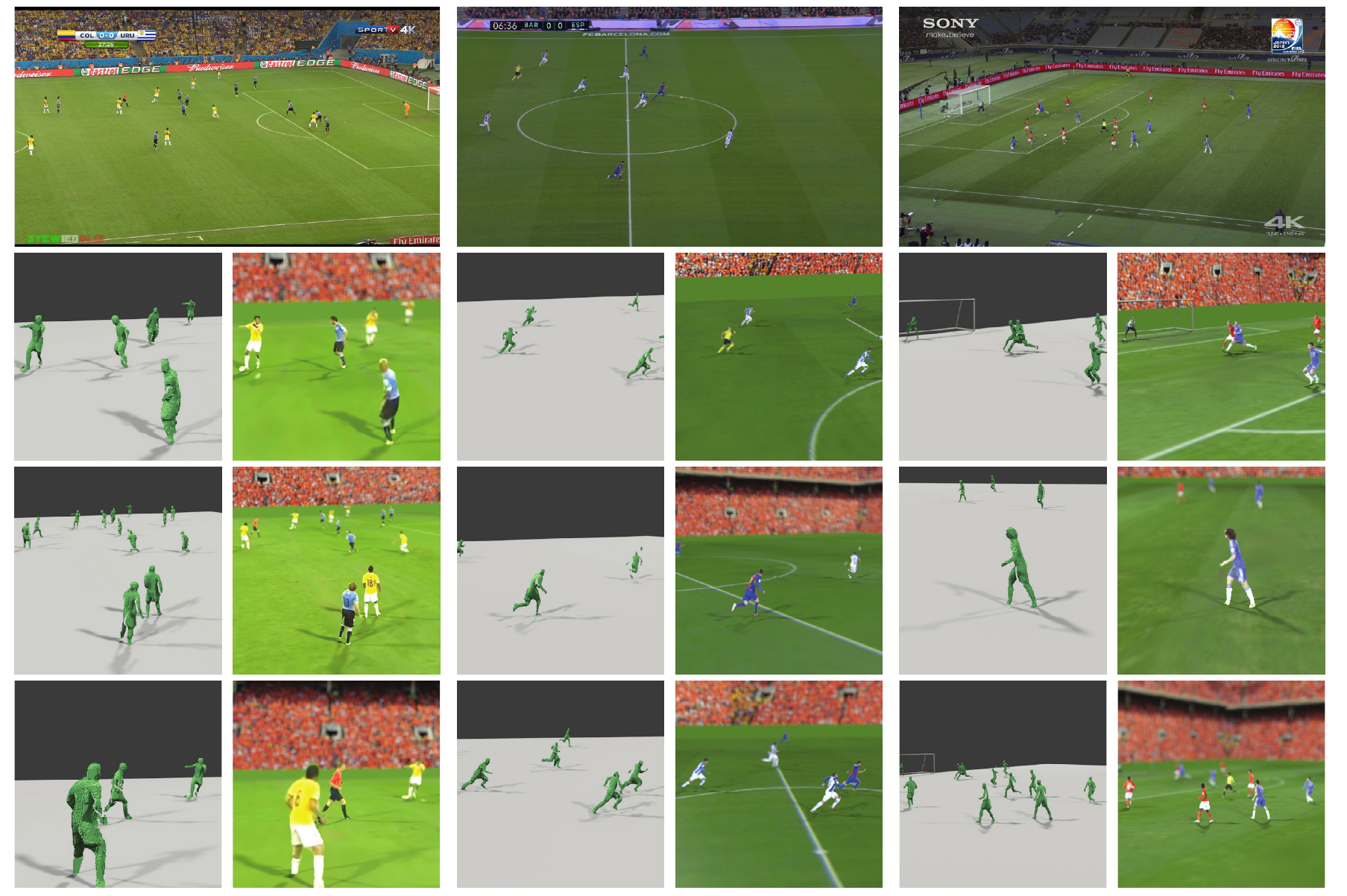}
  \caption{Starting from Youtube frames (top row), the depth maps reconstructed by our network can populate a virtual 3D soccer environment, shown here as mesh-only and textured renderings (rows 2-4).}
  \label{fig:big_results_smallsize}
\end{figure*}

\vspace{-10pt}
\paragraph{Depth Estimation Consistency}
Our network is trained on players from individual frames without explicitly enforcing any temporal or viewpoint coherence. Ideally, the network should give compatible depthmaps for a specific player seen at the same time from different viewpoints. In \refFig{kth_small}, we illustrate the estimated meshes on the KTH multiview soccer dataset~\cite{kazemi2013multi}, with a player captured from three different, synced cameras.
Since we do not have the location of the player on the field, we use a mock-up camera to estimate the 3D bounding box of the player. The meshes were roughly aligned with manual correspondences. 

In addition, for slight changes in body configuration from frame to frame, we expect the depthmap to change accordingly. \refFig{time_consistency_smallsize} shows reconstructed meshes for four consecutive frames, illustrating 3D temporal coherence despite frame-by-frame reconstruction.

\vspace{-10pt}
\paragraph{Experiencing Soccer in 3D} The textured meshes and field we reconstruct can be used to visualize soccer content in 3D. \refFig{big_results_smallsize} illustrates novel views for three input YouTube frames, where the reconstructed players are placed in a virtual stadium.  The 3D video content can also be viewed in an AR device such as a HoloLens (\refFig{teaser}), enabling the experience of watching soccer on your tabletop.  {\bf See supplemental video.}

\vspace{-10pt}
\paragraph{Limitations} 
Our pipeline consists of several steps and each one can introduce errors. Missed detections lead to players not appearing in the final reconstruction. Errors in the pose estimation can result in incorrect trajectories and segmentation masks (\eg missing body parts). While our method can handle occlusions to a certain degree, in many cases the players overlap considerably, causing inaccurate depth estimations.
We do not model jumping players since we assume that they always step on the ground.
Finally, strong motion blur and low image quality can adversely affect the performance of the depth estimation network. 

\mysection{Discussion}{discussion}
We have presented a system to reconstruct a soccer game in 3D from a single YouTube video, and a deployment that enables viewing the game holographically on your tabletop using a Hololens or other Augmented Reality device.
The key contributions of the paper are the end-to-end system and a new state-of-the-art framework for player depth estimation from monocular video.

Going forward there are a number of important directions for future work.  First, only a depth map is reconstructed per player currently, which provides a satisfactory viewing experience from only one side of the field.  Further, occluded portions of players are not reconstructed.  Hallucinating the opposite sides (geometry and texture) and occluded portions of players would enable viewing from any angle.
Second, further improvements in player detection, tracking, and depth estimation will help reduce occasional artifacts and reconstructing the ball in the field will enable a more satisfactory viewing of an entire game. 
In addition, video game data could provide additional information to learn from, \eg, temporal evolution of a player's mesh (if real-time capture is possible using a different capture engine) and jumping poses that could be detected from depth discontinuities between the player and the field.

Finally, to watch a full, live game in a HoloLens, we need both a real-time reconstruction method and a method for efficient data compression and streaming.

\paragraph{Acknowledgements} This work is supported by NSF/Intel Visual and Experimental Computing Award $\#1538618$ and the UW Reality Lab.

{\small
\bibliographystyle{ieee}
\bibliography{egbib}
}

\end{document}